\documentclass[10pt,twocolumn,letterpaper]{article}

\usepackage{wacv}
\usepackage{times}
\usepackage{epsfig}
\usepackage{graphicx}
\usepackage{amsmath}
\usepackage{amssymb}
\usepackage[inline]{enumitem}

\AddToShipoutPicture*{\small \sffamily\raisebox{1.2cm}{\hspace{1.8cm} \textsuperscript{\textcopyright}2020 IEEE. Accepted for publication in: Winter Applications of Computer Vision (WACV) 2020}}

\usepackage[pagebackref=true,breaklinks=true,letterpaper=true,colorlinks,bookmarks=false]{hyperref}

\wacvfinalcopy

\ifwacvfinal\fi
\begin{document}

\title{A Novel Inspection System For Variable Data Printing Using Deep Learning}

\author{Oren Haik \hspace{2cm} Oded Perry \hspace{2cm} Eli Chen \hspace{2cm} Peter Klammer \\
HP Inc.\\
{\small\{\tt haik.oren, odperry, eli.chen.1, pjklammer\}@gmail.com}
}

\maketitle
\ifwacvfinal\thispagestyle{empty}\fi

\begin{abstract}
   We present a novel approach for inspecting variable data prints (VDP) with an ultra-low false alarm rate (0.005\%) and potential applicability to other real-world problems.  The system is based on a comparison between two images: a reference image and an image captured by low-cost scanners. The comparison task is challenging as low-cost imaging systems create artifacts that may erroneously be classified as true (genuine) defects. To address this challenge we introduce two new fusion methods, for change detection applications, which are both fast and efficient. The first is an early fusion method that combines the two input images into a single pseudo-color image. The second, called Change-Detection Single Shot Detector (CD-SSD) leverages the SSD by fusing features in the middle of the network. We demonstrate the effectiveness of the proposed deep learning-based approach with a large dataset from real-world printing scenarios. Finally, we evaluate our models on a different domain of aerial imagery change detection (AICD). Our best method clearly outperforms the state-of-the-art baseline on this dataset.
\end{abstract}

\section{Introduction}
Print defects detection is a necessary step to ensure printing quality. Although manual human inspections are still being employed, automated visual inspection has the potential to replace manual labor due to its accuracy, speed, relative ease of implementation and reduced costs. The key idea in print inspection systems is detecting genuine differences (changes) between a pair of images: a reference image and its corresponding printed (and scanned) image.

Variable data printing (VDP) is a form of digital printing in which elements such
as text, graphics, and images may be changed from one printed piece to the next without stopping or slowing down the printing process \cite{ref58}. VDP inspection is a challenging task as illustrated in Fig. 1 because ‘every page is different’ (e.g., photos) and thus there is high variability in both defect types and image types.
\begin{figure}[!htb]
\begin{center}
\includegraphics[width=0.63\linewidth]{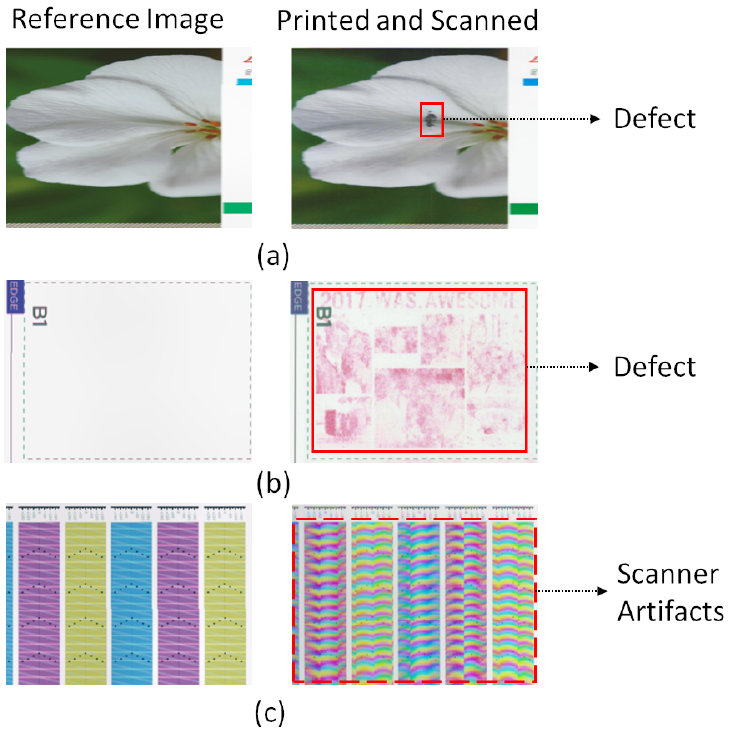}
\end{center}
   \caption{Examples of the main challenges in our system:  (a)	a small defect with respect to the size of the image, (b) a complex defect with high variability  and (c) scanner artifacts that may increase the false alarm rate.}   
\label{fig:short}
\end{figure}
For example, Fig. 1(b) illustrates a complex defect caused by erroneously printing the previous image on top of the current image (`memory' defect). Such defect (change) is difficult to define a priori as there is no limitation on the content of each printed image. 
Also, there are changes of interest called semantic or structural changes (\eg, an appearance of defects as shown in Fig. 1(a) and 1(b)) but also  nuisance changes which are called noisy changes. The noisy changes are caused by multiple variables like changes in illumination, misregistration and low-cost scanner artifacts (Fig. 1(c)) that can be hard to isolate from the structural changes. Any detection system must be able to robustly differentiate between real defects (semantic/structural changes we care about) and noise (e.g., due to scanner artifacts).

A defect detection process can be framed as either an object detection task or a segmentation task	\cite{ref22,ref4}. Fully convolutional networks (FCN) have dominated the recent progress \cite{ref22,ref3}. However, the vast majority of deep learning (DL) based inspection systems take as input only the potential defective image without the corresponding reference image	\cite{ref23,ref22,ref25,ref24,ref3}.
Single-frame object detection methods such as Faster R-CNN \cite{ref5} and the single shot detector (SSD) \cite{ref13} are commonly used for detection tasks while Mask-RCNN \cite{ref26}, U-net \cite{ref27} (inspired) or DeepLab v3+ \cite{ref55} networks are used for semantic/instance segmentation (prediction at the pixel level). There is no need for a reference image in such systems as the focus is on detecting specific defect types per application. It is also assumed that the defect instances in each class are quite homogeneous (including the background on which each defect is overlaid). Thus, each method has its own characteristics that only respond to specific kinds of features. This is in contrast to the VDP inspection system that should detect a wide range of defect types (instead of limited and specific `changes') that do not necessarily appear in the training set. In addition, each print is potentially different; thus, a reference image must be generated for each.

Given a pair of images, change detection is the most related domain for VDP inspection systems. Such techniques \cite{ref32,ref20,ref21,ref30,ref29,ref28,ref53,ref33, ref31,ref4,ref34} use an aligned image pair as an input and return either a pixel-wise classification map of the structural changes (semantic segmentation) or a bounding box around each changed region. They can be classified into three categories depending on the stage at which the two images are fused \cite{ref32,ref21,ref28,ref42}: pixel level (early fusion), feature level (medium fusion) and decision level (late fusion). 

In this paper, we propose a novel early fusion method, which is both fast and efficient. It significantly outperforms the common baselines of combining the two RGB images into a single image via concatenation (along the channel dimension) or by a difference \cite{ref32,ref31,ref54}.

We also propose a novel feature level (medium) fusion method for detecting genuine differences between two images using a variant of the SSD. It is based on an efficient Siamese architecture that merges the data in a way that is the most appropriate for our application. We show that it significantly outperforms the recently proposed Siamese SSD architecture by V. Osin \etal \cite{ref35}.

Our main contributions are summarized as follows :
\begin{itemize}
  \item To the best of our knowledge, we are the first to introduce an automated, end-to-end, real-time (production speed - at least one page per second), industrial and low-cost inspection system for VDP, while still maintaining low false alarm and miss detect rates.  
  \item We propose a novel early fusion method, for change detection applications, which is both fast and efficient. It is based on combining the two input RGB images (reference and scanned images in our application) into a single pseudo-color image that enhances the semantic changes while preserving the essential image pair information. This image can be used as an input to any single-frame object detection methods (\eg, SSD) including pre-trained models (transfer learning).
\item We present a novel Siamese network architecture, called Change-Detection SSD (CD-SSD), for detecting semantic changes between two images. It is based on a network design that leverages the SSD efficiently by fusing features in the middle of the network. As far as we know, this is the first time
a SSD based network has been applied for change detection applications. 
\item Our methods outperform the baseline methods by a large margin when evaluated on a large dataset of real defects from real-world printing scenarios. We also demonstrate the potential of applying our methods, in other areas, by training each on the publicly available Aerial Imagery Change Detection (AICD) \cite{ref52} dataset. Our best model (CD-SSD) clearly outperforms the state-of-the-art solution on this dataset. 
\end{itemize}

\section{Related Work}
\textbf{Classical computer vision based techniques.}
Traditional computer vision-based inspection systems have been relatively well-studied. These systems occur in many industrial applications \cite{ref51,ref36,ref37} like printed circuit boards, textile and texture inspection. 
One popular method is subtracting a reference image from a potentially defective image and then thresholding the result. Although this method is fast, it tends to be very sensitive to noisy changes (such as misregistration errors). Thus, it does not work well without significant pre/post-processing \cite{ref53}. The same is true when using more advanced `classical' image quality metrics such as the structural similarity index measure (SSIM) \cite{ref38} or even DL based image quality measures \cite{ref1}. To address these limitations, some methods have proposed to extract features from each defect candidate. The features are then put into a classifier that is trained in advance to determine whether it contains a defect or not. Hand-crafted features \cite{ref51,ref39} or DL based features \cite{ref2} are usually used for this task. However, the extracted features tend to be application specific and don't solve the defect localization problem. 

\textbf{Single frame (no-reference) DL based networks.}
With the recent success of deep neural networks on generic object detection (SSD, Faster R-CNN, \etc) and segmentation (Mask-RCNN, U-net, \etc) networks it becomes very natural to use them as a basis for fast and accurate defect detection systems \cite{ref60,ref57}. Application examples are printed circuit boards \cite{ref23}, railway track inspection \cite{ref24}, metallic surface detection \cite{ref3}, sealing surface inspection \cite{ref25} and casting defects in X-ray images \cite{ref22}. However, 
such methods are not applicable to the case of VDP because they are focused on detecting application-specific defects, whereas the VDP problem requires a 'general' defect detector that can be extended well to new (previously unseen) defect types

\textbf{Change detection using semantic segmentation.}
Much research has been done in the field of 
change detection. The core idea in DL-based methods is to fuse the two images \cite{ref32,ref21,ref28,ref42} using early, medium, or late fusion and then use a FCN to predict per-pixel segmentation map.

In the case of early fusion, it is common to concatenate the two RGB images into a single 6-channel image which is used as an input to a FCN \cite{ref32,ref20,ref30,ref31}. However, as the filters of the first convolutional layer are modified (due to using 6 channels instead of 3) single frame pre-trained models cannot be reused directly. A workaround is replicating the weights of the first layer along the channel dimension (similarly to bootstrapping 3D filters from 2D filters \cite{ref41}) and use it as an initialization method. It requires one to fine-tune the lowest layer (or few lowest layers) in the model which is risky in case of a small dataset. A difference between the two images \cite{ref31} is another option, but it is more sensitive to noise and misregistration errors. 

Different from early fusion, middle fusion  is based on merging middle-level convolutional features \cite{ref32,ref21,ref35,ref33,ref34}. It usually provides better results as it fuses information at a stage where the spatial features are less relevant. However, it is unclear a priori which architecture of the fusion model could get the best results for a specific application \cite{ref54}. This includes the layers to be merged and the data fusion functions (concatenation, difference, max, \etc).

Decision level fusion follows a two-stage approach \cite{ref28,ref42} - each image is first passed independently through a FCN for predicting a binary segmentation mask. Then the binary masks are subtracted. However, it assumes low variability between instances of the same class.

The main disadvantage of semantic segmentation networks is a high computational cost (mainly for large images). The need for a large amount of per-pixel labeled data is another disadvantage. Some methods alleviate this problem by using lightweight network architectures \cite{ref20,ref44}, but the accuracy is usually lower. Also, these networks are typically trained with the cross-entropy loss function \cite{ref20}, or the contrastive loss function \cite{ref21} that essentially makes the network learn a classifier for each pixel (and summing up the loss). This makes it hard to train the network when the objects to detect are small \cite{ref45,ref43}. Finally, the training and evaluation are usually done on datasets with limited types of changes (appearing/disappearing of buildings, construction areas, vehicles, vegetation, \etc). Such task-specific networks may not generalize well to `general' change detection applications like VDP. 

\indent 
\textbf{Multi-frame object detection networks.}
Object detection is an efficient and faster alternative to semantic segmentation.
In the case when multiple frames are available, the additional temporal/depth information can be used for improving detection accuracy. There are many different ways to model motion cues, including 3D convolutional neural networks or recurrent neural networks \cite{ref47,ref48,ref41,ref46}. Two-stream based networks are the most common approach \cite{ref46}. It consists of a spatial network that models appearance with RGB frames as an input and a temporal network that models motion (optical flow). Then, decision level fusion, which works on the bounding-box level is commonly used.  However, the focus of such methods is not on detecting changes between two frames but rather on improving per-object detection accuracy (compared to single-frame models), object tracking and action recognition. 

Recently, V. Osin \etal \cite{ref35} extended the SSD to support multi-spectral data (visible and infra-red images). It has a Siamese network architecture that fuses the two branches at each detection layer of the SSD. The data fusion function is a concatenation followed by 1$\times$1 convolution filters. We show later (Section 5) that for VDP, it falls below our proposed Change-Detection SSD network (CD-SSD) by a large margin (in terms of accuracy). 

J. Wu \etal \cite{ref4} proposed to consider the differences between two (book cover) images (captured with a high-resolution camera) as the objects to detect. The two RGB images are concatenated (early fusion) into a single 6-channel image. Then, Faster R-CNN \cite{ref5} is used to spot the difference between them. We show later (Section 5) that our novel early fusion method (`Pseudo-color') significantly outperforms it.

\section{Legacy Inspection System}

Our first-generation system was designed using classical computer vision based techniques. A block diagram of the system is shown in Fig. 2. It is based on \cite{ref56} and contains the following stages:
\begin{figure*}[!htb]
\begin{center}
\includegraphics[width=0.75\linewidth]{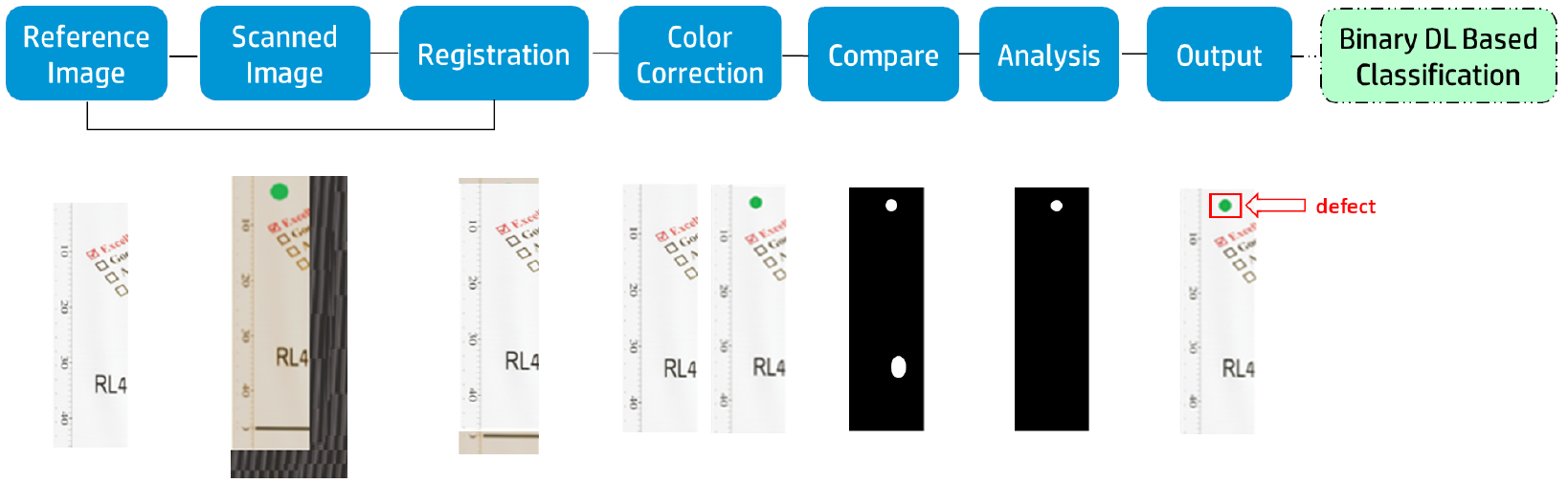}
\end{center}
   \caption{A block diagram of the legacy system. The two input images (reference image and its printed and scanned version) are registered and color corrected before comparing them using the SSIM. False alarms can be reduced by applying a binary  classifier (true/false defect) using hand-crafted/deep learning (DL) based features.}   
\label{fig:short}
\end{figure*}

\subsection{Pre-Processing}
\textbf{Creating a reference image (RGB).}
The reference and scanned images are in different color spaces. A lookup table is used to convert the reference image from CMYK to  RGB (scanner) color space. After this stage, the two images are almost color matched and in the same RGB color space.

\textbf{Page corners detection.} 
Before applying the registration filter, the paper image should be cropped from the scanned image which contains scanner background (stripes pattern of the conveyor belt \textendash see Fig. 2) followed by rotating it into zero angle orientation. The page's corners are detected by first identifying discontinuities in the standard deviations of gray levels along rows and columns in the scanned image. Then the Harris corner detector \cite{ref8} is applied but only in a small region around each discontinuity. 

\subsection{Registration And Color Correction}

The reference image should be aligned on a pixel-by-pixel basis with respect to the scanned image. This is because paper movement causes local and global spatial distortions. A global template matching (block matching) \cite{ref9} for coarse alignment between the images is followed by a local template matching for fixing the local movements. 
Histogram match \cite{ref10} is used for reducing color inconsistencies (caused during scanning) between the two images.

\subsection{Image Comparison (SSIM)}

The registered images are compared using the SSIM metric of Z. Wang \etal \cite{ref38}, patch versus patch. Each patch in the scanned image receives a score that represents its similarity to the reference patch. In order to improve the detection rate of mainly large defects and increase robustness to small misregistration errors, the SSIM is also applied on several down-sampled versions of the scanned and reference images (multi-scale SSIM). 

\subsection{Analysis}

Every defect that is found in the SSIM map (after thresholding) is analyzed, to reduce false alarms caused by small misalignment errors or scanner artifacts such as: moiré (under-sampling of fast color variation- example in Fig. 1(c)), dust (dirty scanner), noise and illumination inconsistency (un-calibrated scanner). This was done by extracting simple shape and texture features (size, contrast, \etc) and filtering the irrelevant defects using empirical thresholds (\eg, minimal defect size is about 1 mm squared). 

\subsection{Output}

In order to get the full defect area, region growing \cite{ref10} is applied to each detected defect. Reference and scanned images are presented to the user with red rectangles around each defect. Binary DL based classification network (\eg, VGG16), pre-trained on the ImageNet dataset, may be applied on each defect candidate for reducing false alarms.

\section{Proposed Method}
Inspired by the recent success of deep convolutional neural networks in the field of object detection, we explore two options to leverage it for detecting genuine differences (print defects) between two images.  A block diagram of the proposed DL based system is presented in Fig. 3. It comprises the same pre-processing module (Section 3.1) as the legacy system. Then, FlowNet2 \cite{ref12} is used for optical flow computation (registration). It is faster and more accurate (maximum error - few pixels) compared to local template matching (Section 3.2). The printed page is stretched randomly (and locally) due to the media transport mechanism of our system, which is complex and non-ideal. Therefore a non-global transformation (like FlowNet2 or PWC \cite{ref49}) is needed to estimate local movements between the scanned and reference images. Finally, the two images are fused either using a novel early fusion method called Pseudo-color SSD or a middle-fusion method called Change-Detection SSD (CD-SSD). Both are described next. Meaningful (structural) differences are considered as the objects to detect. Noisy changes (color deviations, scanner artifacts, \etc) are excluded automatically in the training due to the hard-negative mining employed in the SSD. The resultant network learns only the real defects characteristics.

\begin{figure*}[!htb]
\begin{center}
\includegraphics[width=0.59\linewidth]{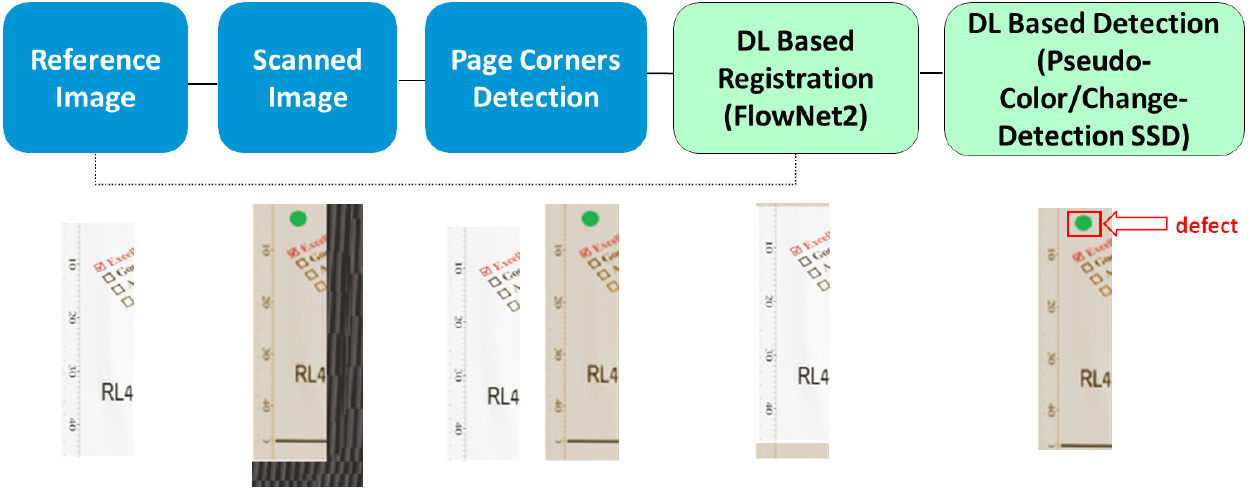}
\end{center}
   \caption{A block diagram of the proposed deep learning (DL) based inspection system. Only the pre-processing module is the same as the legacy system. The registration is based on FlowNet2 and the detection on novel single shot detector (SSD) based methods (Section 4).}
\label{fig:short}
\end{figure*}

\begin{figure}[!htb]
\begin{center}
\includegraphics[width=1.00\linewidth]{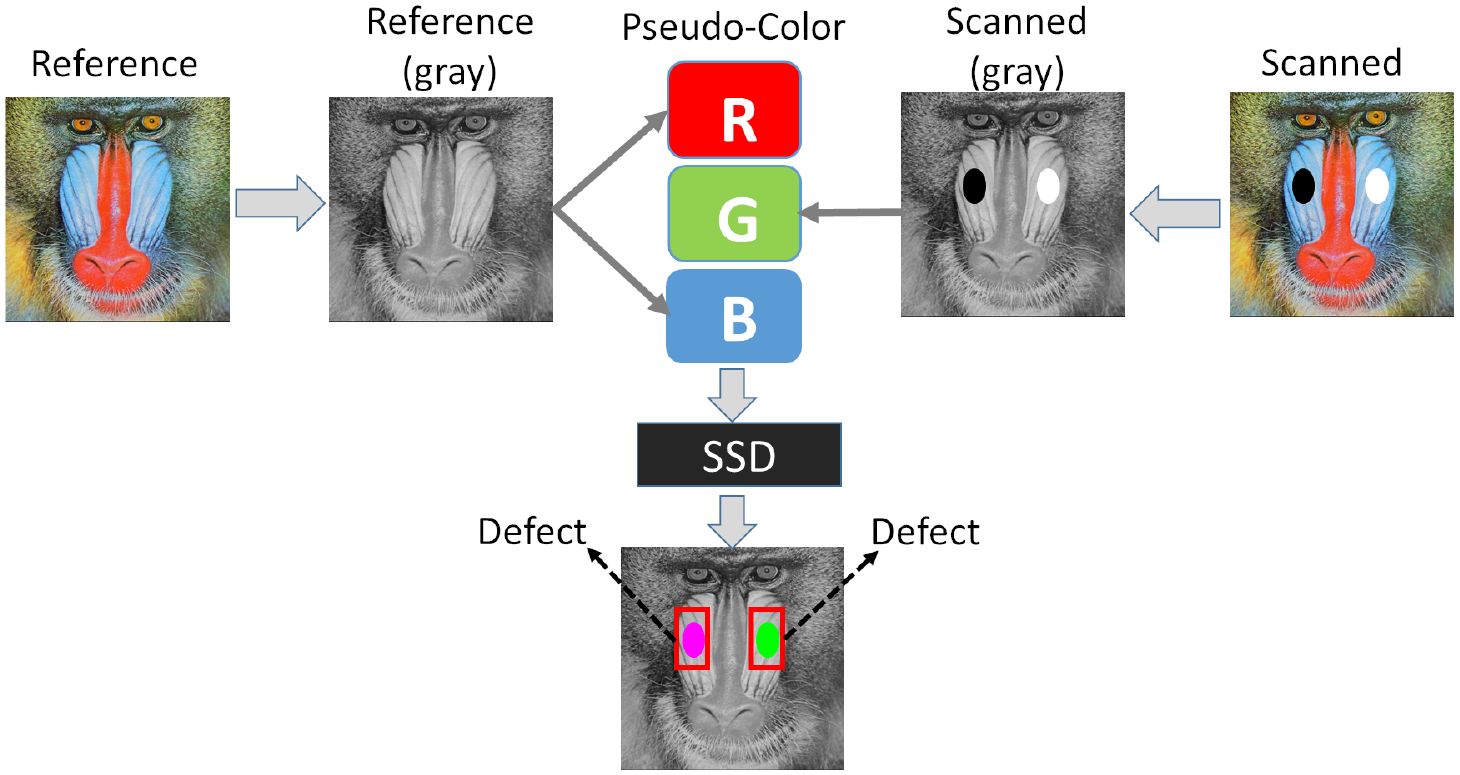}
\end{center}
   \caption{The proposed early fusion method. The reference and scanned images are combined into a single pseudo-color image which is used as an input to the SSD. Grayscale (in the pseudo-color image) denotes no difference between the two images while green or magenta colors denote a difference.}
\label{fig:short}
\end{figure}

\subsection{Pseudo-Color SSD }

Single-frame object detection models (\eg, SSD) expect one input image with three channels while we have a total of six channels (three from the scanned image and three from the reference image). A possible solution is to combine the two images (scanned and reference) into a single image (early fusion). 
Fig. 4 presents our proposed early fusion method. The reference and scanned images are compressed from three channels to one by converting each to a gray-scale image. Then, the scanned gray-scale image is mapped to the green channel of the combined image.   Similarly, the reference gray-scale image is mapped to the red and blue channels of the combined image. Regions in the combined image where the reference and scanned images are identical will appear in grayscale. In contrast, regions in the combined image where the reference image differs from the scanned image will have a green or magenta appearance. In this way, the combined image may be considered to be a pseudo-color image, as the true (i.e., RGB) colors of the reference/scanned image are not apparent. The pseudo-color image can be used as an input to the SSD. The intuition is as follows:

\begin{figure*}[!htb]
\begin{center}
\includegraphics[width=0.64\linewidth]{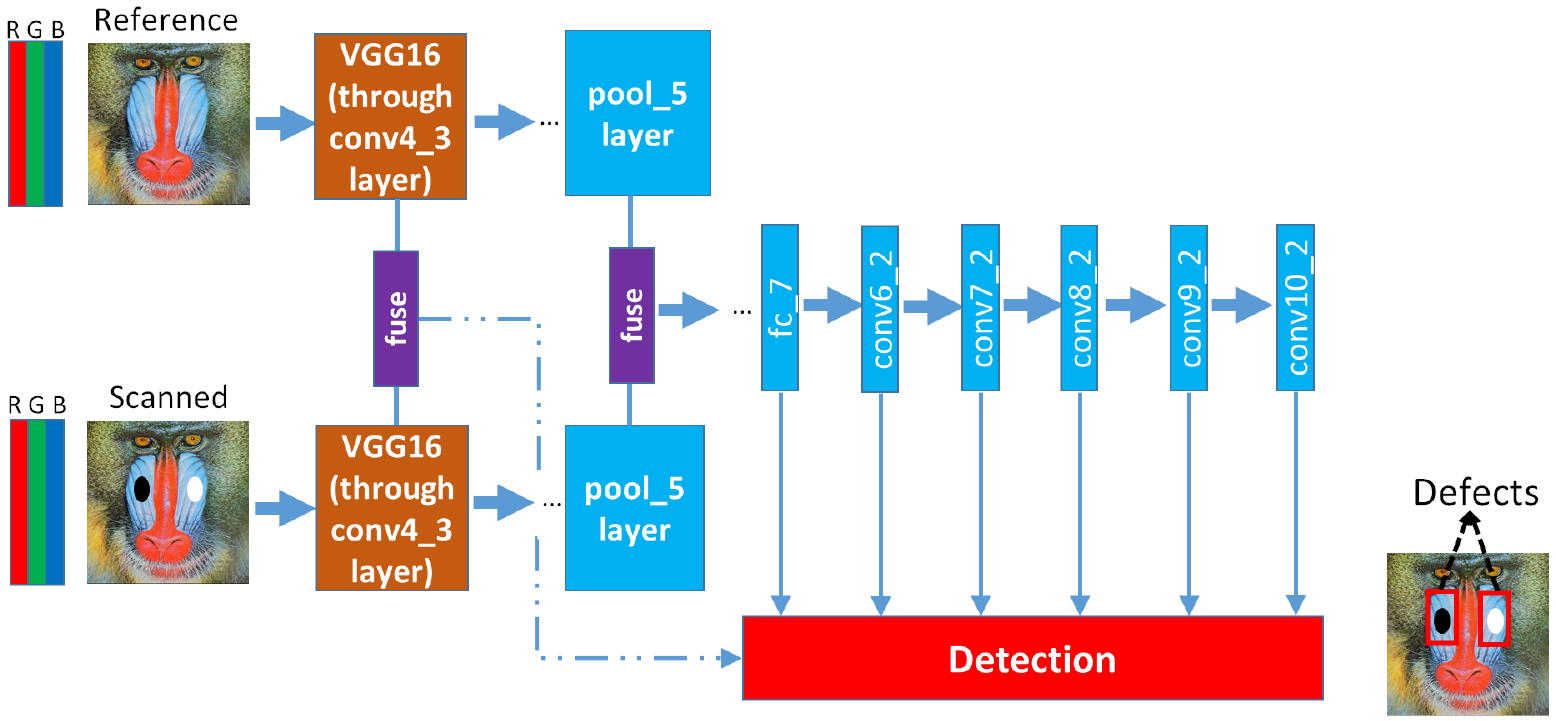}
\end{center}
   \caption{Illustration of the proposed Change-Detection SSD (CD-SSD) architecture (middle fusion). The two input images are fused twice (using a concatenation) after conv4\_3 and pool\_5 layers of VGG16. Next, the architecture is the same as the single frame SSD (512 model).}
\label{fig:short}
\end{figure*}
\begin{itemize}

\item
In case of VDP, what constitutes a real defect compared to a false alarm is mainly a local change in the structure of the scene and not its color (as it may appear in various colors). Such structural/semantic changes are clearly evident in the grayscale version of the images (except very low contrast differences).

\item 
Concatenating the two gray-scale images (along the channel dimension of the pseudo-color image) is more robust to misregistration errors compared to taking the difference between them.
\item 
The human eye is more sensitive to the green color. This means better enhancement of defects when mapping the scanned image (gray-scale) into the green-channel of the pseudo-color image. Note that there is no need, in our system, to detect defects that are not visible to the human's eyes (\eg, due to the background on which each defect is overlaid).

\item 
Replicating the reference image (gray-scale) results in a 3-channel image (two from the reference and one from the scanned). This pseudo-color image has many features that qualitatively appear in standard RGB images, such as edges and corners (see Figs. 4 and 7). Therefore, using single-frame pre-trained models (\eg, on the ImageNet dataset) is possible in this approach. This is in contrast to the common early fusion (concatenate) method of using a 6-channel image.
\end{itemize}

\subsection{Change-Detection SSD (CD-SSD)}

The architecture of CD-SSD is shown in Fig. 5. It has two main components:  a feature extractor network (VGG16) and a detection `head' consisting of convolutional layers. The main modification we make to the SSD512 architecture is adding two data fusion (concatenation) layers after conv4\_3 and pool\_5 layers of VGG16 (the weights of the two branches are shared before the fusion layers). The output of the data fusion layer after conv4\_3 is fed into the first detection layer of the SSD. The architecture after pool\_5 fusion layer is the same as a standard single-frame SSD512 model. The intuition is as follows:
\begin{itemize}
\item The first fusion is at conv4\_3 layer (and not before) since it has a relatively large receptive field which is more robust to misregistration errors.
\item 
We use two fusion layers (instead of one) in order to exploit  ImageNet pre-trained weights until pool\_5 layer of VGG16. This is in contrast to V. Osin \etal \cite{ref35} which use seven fusion layers in case of SSD512 (one for each detection layer).    
\item 
Feature maps concatenation is used instead of convolution fusion \cite{ref35} which adds 1$\times$1 convolution filters on top of it (for dimensionality reduction). This is because 1$\times$1 convolutions are less robust to misregistration errors and also may lose some data (the same is true for fusion using a difference).
\end{itemize}


\begin{table*}[!htb]
\begin{center}
\begin{tabular}{|c|c|c|c|c|}
\hline
Method &  False Alarms  &  Miss Detect  & Execution time & Number of \\
 & (\%)& (\%) & per sample (sec) & trainable parameters (M) \\
\hline\hline
Legacy & 0.5 & 30 &0.90 & \textbf{0}  \\
Legacy + Classification (VGG16) & 0.050 & 32 &1.00 & 134.265\\
FlowNet2 + Pseudo-color SSD (ours) & 0.010  & 15  &0.25 & 24.386  \\
FlowNet2 + CD-SSD (ours)  & \textbf{0.005}  & \textbf{10} & 0.30 & 29.216 \\
\hline
FlowNet2 + Concatenate + SSD & 0.020  & 35  &0.25 & 24.388 \\
FlowNet2 + Multi-spectral SSD \cite{ref35} & 0.015 & 23 & 0.30 & 28.059\\
FlowNet2 + DIFF + SSD & 30  & 22  &0.25 & 24.386 \\
\hline
SSD (scanned image) \cite{ref13} & 55  & 48  &\textbf{0.07} & 24.386 \\

\hline
\end{tabular}
\end{center}
\caption{Detection results evaluated on the real defects (labeled) dataset of 40,000 defects using a 4-fold cross-validation. We compare our detection models (lines 1-4) with four baseline methods (lines 5-8).}  
\end{table*}

\begin{table}[!htb]
\begin{center}
\begin{tabular}{|c|c|}
\hline
Method &  False Alarms \\ & (\%) \\
\hline\hline
Legacy &0.5      \\
Legacy + Classification (VGG16) & 0.1 \\
FlowNet2 + Pseudo-color SSD (ours) & 0.010    \\
FlowNet2 + CD-SSD (ours)  & \textbf{0.005} \\
\hline
FlowNet2 + Concatenate + SSD & 0.025 \\
FlowNet2 + Multi-spectral SSD \cite{ref35} & 0.020   \\
FlowNet2 + DIFF + SSD & 35   \\
\hline
SSD (scanned image) \cite{ref13}  & 60  \\

\hline
\end{tabular}
\end{center}
\caption{Detection results (in terms of false alarms) evaluated on a test set of 80,000 image pairs from real-world printing scenarios.}
\end{table}

\vspace{-2.0ex}%
\section{Experiments}

\subsection{Datasets}
  
Two datasets from real-world printing scenarios were used to train/evaluate our models. The datasets are an order of magnitude larger than existing change-detection datasets and much more challenging. This is due to high variability in image/defect types (size, shape, contrast, \etc), illumination changes, misregistration and low-cost scanner artifacts. Fig. 6 contains examples of some defects from the datasets, cropped from the full scanned images (which contain   500 $\times$ 1250 pixels). 

\begin{figure}[!htb]
\begin{center}
\includegraphics[width=0.50\linewidth]{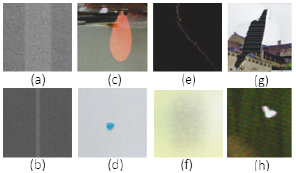}
\end{center}
   \caption{Examples of some defects from our datasets :  (a) band, (b) streak, (c) large drip, (d) small drip, (e) paper wrinkle, (f) random spots, (g) paper cut and (h) dent.}
\label{fig:short}
\end{figure}

\noindent \textbf{Real defects dataset (labeled)} -
A dataset of 20,000 image pairs with 40,000 real defects. The labeling, meaning drawing bounding boxes around the true defects, was done using the LabelImg tool \cite{ref18}. The dataset was divided into training and validation sets using a 4-fold cross-validation.
\newline 
\noindent \textbf{Real defects test-set (unlabeled)} -
A test set of 80,000 image pairs. Some of the images are defect free, and some may contain more than one defect. The set is unlabeled which means that we can measure only the false alarm rate.  False alarm rate is most important in high speed press application, since high rates of false alarm (even 0.1\%) will become annoying to the operator/user causing them to ignore or disable the inspection system.

\subsection{Evaluation Metrics}

False alarm rate (FPR) and miss detect rate (FNR)\qquad  are used to evaluate the accuracy of each method: 
\begin{equation} \label{eq1}
\begin{split}
FPR &= \frac{FP}{TP+FP}\times100\thinspace[\%]\\
\end{split}
\end{equation}
\begin{equation} \label{eq1}
\begin{split}
 FNR &= \frac{FN}{TP+FN}\times100\thinspace[\%]\\
\end{split}
\end{equation}

\noindent where FN are the false negatives and FP are the false positives. Detection is a true positive (TP) if the Intersection Over Union (IOU) with the ground-truth box is above 0.25.  
\subsection{Implementation Details}

The training was done with the real defects (labeled) dataset, Adam optimizer, a learning rate of 0.0001 with linear decay rate \cite{ref59}, 200 epochs, batch size of 8, one class (real defects), MS COCO scales and input images resized to 512$\times$512 pixels. The other parameters were set to their default values according to the original SSD paper \cite{ref13}. The layers of the base network (VGG16 until pool\_5 layer) were frozen during the first 100 epochs followed by unfreezing them in the next 100 epochs. They were initialized with ImageNet pre-trained weights, while the rest of the layers were trained from scratch (Xavier initialization). The code is based on the publicly available Keras SSD512 implementation \cite{ref19}.  We performed a 4-fold cross-validation and averaged the results. The following on the fly augmentations were employed (randomly): horizontal/vertical flips, color channels swap and contrast/brightness stretch. NVIDIA Quadro P5000 GPU was used for training and testing.

\begin{figure*}[!htb]
\begin{center}
   \includegraphics[width=0.76\linewidth]{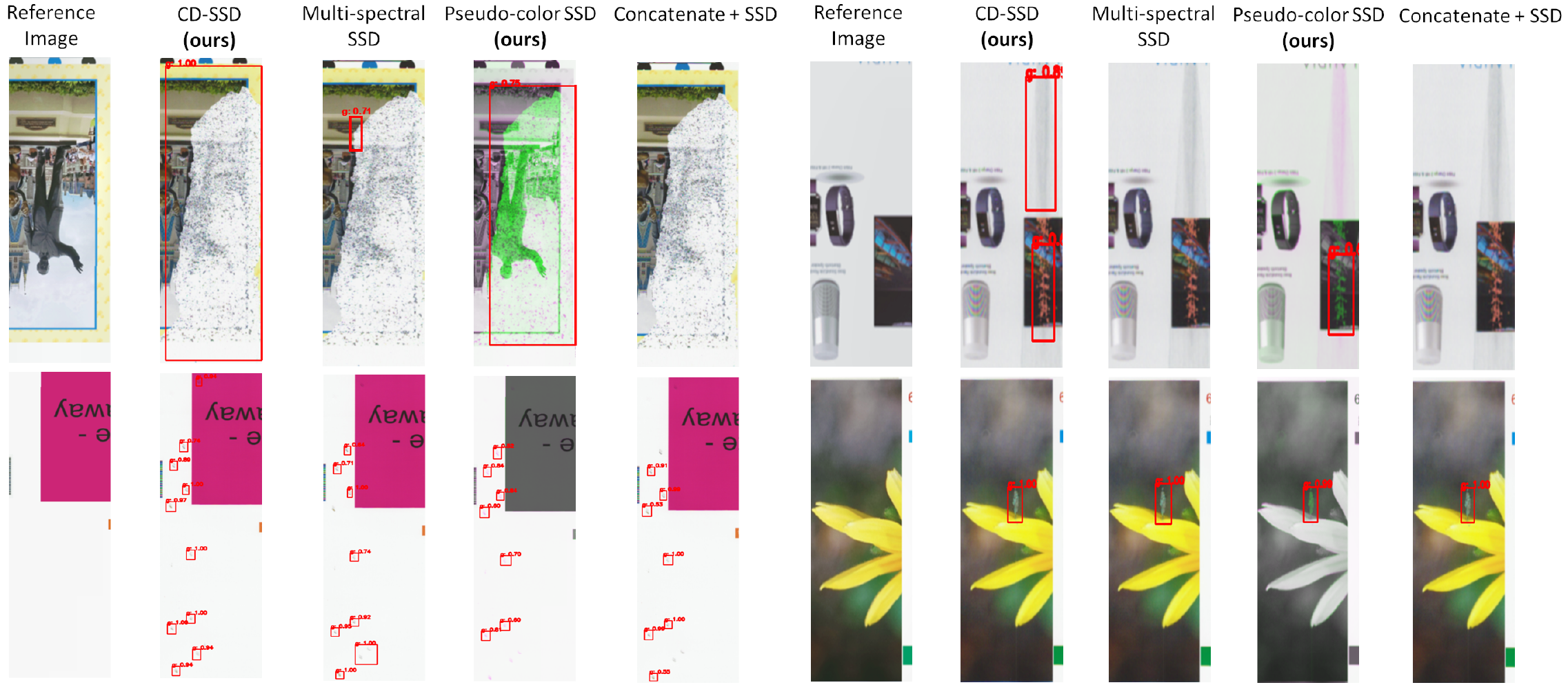}
\end{center}
   \caption{Some qualitative results (red bounding boxes around the defects) from one of the validation sets. One can note that our methods perform well on defects with diverse properties (size, shape, contrast, number of defects in the page, \etc) compared to the baseline methods.}
\label{fig:long}
\label{fig:onecol}
\end{figure*}

\subsection{Comparative Analysis}
We compare our methods (Section 4) with various baselines which include: early fusion using a concatenation or a difference (DIFF) between the two images, multi-spectral SSD \cite{ref35}, single frame SSD \cite{ref13} (only scanned image) and our legacy system (Section 3). The training of all the SSD based models follows the same procedure described above (Section 5.3). In the case of an early fusion using concatenation, the filters of the first convolutional layer (6 channels) were initialized by replicating ImageNet first layer (3 channels) weights along the channel dimension. This allows using transfer learning in this case. We did experiments with many fusion options, but we report (in Table 1) only the schemes which provided the best results. Object detection based solutions are used as baselines and not semantic segmentation (Section 2) based methods because most of them are not fast enough for our application. In order to inspect every printed page, the execution time of all the modules (pre-processing, registration, detection) must be less than one second (a typical production speed of a printing press). Some methods use a large output stride (\eg, 16 in \cite{ref55}) to reduce the computational cost, but this means less detailed segmentation map. It is a disadvantage in our case, as some of the defects are very small (a few pixels- \eg, in Fig. 7). State of the art multi-frame object detection methods use more than two images (a video) as an input, so they are not applicable in our case as well.  
\subsection{Results}
For the real defects (labeled) dataset we report in Table 1 the accuracy (FPR and FNR), execution time and the number of trainable parameters attained by each method. It can be noted that:
\begin{enumerate*}[label={\alph*)},font={\bfseries}]
\item CD-SSD and Pseudo-color SSD outperform all baselines by a large margin including the legacy system (in terms of accuracy).
 \item CD-SSD is better than `Pseudo-color' but with a relatively small margin (accuracy). This means that when performance limitations (speed/memory) are an issue, the `Pseudo-color' could be a good alternative. It allows using any single-frame DL based object detection model `as is' including transfer learning. Although theoretically, a CNN with more than three channels input (Table 1, line 5) could have learned a similar/better color encoding parameters it failed to do so in case of data with high variability. 
\item Training the SSD without a reference image (only scanned) or with a difference (DIFF) between the two images completely failed (very high false alarm rate). This is expected since in VDP each print is potentially different; therefore, a reference image must be generated for each. Also, a difference is more sensitive to noise (\eg, scanner artifacts) and misregistration errors.
\item The specific two-branch network design (CD-SSD vs. Multi-spectral SSD) affects the FPR and FNR significantly. It includes the number and location of the fusion layers and also the fusion function (\eg, concatenate/convolution \cite{ref35}).
\end{enumerate*} 

Fig. 7 presents some examples (validation set). We can see that our methods (in contrast to the best baselines) detect well a wide variety of defect types with different size, shape, contrast, \etc 

Fig. 8 contains examples that demonstrate the importance of a reference image in VDP and also the challenges compared to other computer vision problems. The defects (detected using CD-SSD) in Fig. 8(a) and 8(b) resemble true object parts, tie and door knob, respectively. Therefore without a reference image, it would be very difficult to know whether these are defects or not. Some of the defects are with low contrast (Fig. 8(c)), difficult to define a priori (Fig. 8(d)), and should be detected while ignoring scanner artifacts (Fig. 8(e)).
Table 2 presents for each method the false alarm rate (FPR) evaluated on the real defects test set of 80,000 image pairs. The results are quite similar to those in Table 1.

\begin{figure}[!htb]
\begin{center}

\includegraphics[width=0.85\linewidth]{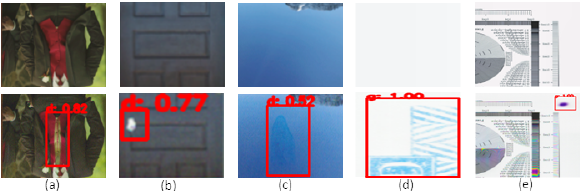}
\end{center}
   \caption{Examples of reference (first row) and scanned (second row) patches: (a) a drip defect that looks like a tie, (b) a dent defect that resembles a door knob, (c) a low contrast defect (drip), (d) a complex defect type with high variability and (e) detection of a real defect while ignoring scanner artifacts. All of the defects (red bounding boxes around them) were detected using CD-SSD.}
\label{fig:long}
\label{fig:onecol}
\end{figure}

\subsection{Results For Different Application}
We trained our models (without the pre-processing module - Section 3.1) on a different application: Aerial Imagery Change Detection (AICD). The publicly available dataset \cite{ref52} contains 1,000 image pairs with variations between them (an example is in Fig. 9). From Table 3, one can see that our CD-SSD method outperforms the state-of-the-art solution on this dataset. Also, in contrast to our datasets, the difference between the single-frame SSD baseline and the other methods is quite small. This emphasizes the challenge of high variability in VDP compared to other change detection applications. Further, as the `Pseudo-color' lacks the `true' color information, it is somewhat inferior to the single-frame baseline. It means that for data with low variability (contrary to VDP) color is an important feature.

\begin{table}[!htb]
\begin{center}
\begin{tabular}{|c|c|}
\hline
Method & Average Precision (\%) \\
\hline\hline
FlowNet2 + Pseudo-color SSD & 96.07\\
S. H. Khan \etal \cite{ref40} & 97.30  \\
FlowNet2 + CD-SSD & \textbf{99.00} \\
\hline
SSD (no reference image) \cite{ref13} & 97.01   \\
\hline

\end{tabular}
\end{center}
\caption{Detection results for the AICD dataset (test set: 30\% of the data). Our CD-SSD clearly outperforms the baseline networks.}
\end{table}

\begin{figure}[!htb]
\begin{center}
\includegraphics[width=0.71\linewidth]{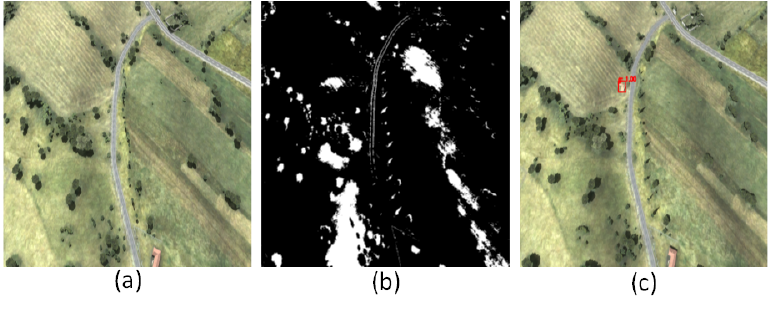}
\end{center}
   \caption{Example of a reference image (a) from the AICD dataset \cite{ref52}, a test
image (c) with our CD-SSD detection (red bounding box) and the thresholded image
difference (b).}
\label{fig:long}
\label{fig:onecol}
\end{figure}

\section{Conclusion}
We presented a novel inspection system for
Variable Data Printing (VDP). It allows one to detect general purpose defects (i.e., without assuming any specific type a priori) with a very low false alarm rate. This was achieved by proposing two new fusion methods, which are both fast and efficient. The first is an early fusion method called `Pseudo-color' and the second is a middle fusion method called Change-Detection Single Shot Detector (CD-SSD). Experiments on large datasets from real-world printing scenarios demonstrate that both methods outperform the baselines by a large margin. Also, we trained our models on the aerial imagery change detection (AICD) dataset, and CD-SSD clearly outperforms the state-of-the-art baseline. Thus, our solution can be applied to other change detection applications.


{\small
\bibliographystyle{ieee}
\bibliography{VDP_ref}
}

\end{document}